# Unsupervised Time-Series Signal Analysis with Autoencoders and Vision Transformers: A Review of Architectures and Applications

**Hossein Ahmadi, Sajjad Emdadi Mahdimahalleh*, Arman Farahat, Banafsheh Saffari**

Department of Electrical and Computer Engineering, The University of Akron, Akron, OH, USA
Email: *se67@uakron.edu

## Abstract

The rapid growth of unlabeled time-series data in domains such as wireless communications, radar, biomedical engineering, and the Internet of Things (IoT) has driven advancements in unsupervised learning. This review synthesizes recent progress in applying autoencoders and vision transformers for unsupervised signal analysis, focusing on their architectures, applications, and emerging trends. We explore how these models enable feature extraction, anomaly detection, and classification across diverse signal types, including electrocardiograms, radar waveforms, and IoT sensor data. The review highlights the strengths of hybrid architectures and self-supervised learning, while identifying challenges in interpretability, scalability, and domain generalization. By bridging methodological innovations and practical applications, this work offers a roadmap for developing robust, adaptive models for signal intelligence.

## 1. Introduction

With the increasing volume of unlabeled time-series data in domains such as wireless communications, radar systems, biomedical engineering, and the Internet of Things (IoT), the need for robust unsupervised learning methods has become more pronounced. Traditional supervised models—including CNNs and RNNs—require large labeled datasets, which are often expensive or impractical to collect in real-world environments. In contrast, unsupervised learning models, particularly Autoencoders (AEs) and Vision Transformers (ViTs), offer flexible alternatives for extracting meaningful representations, detecting anomalies, and classifying signals without reliance on manual labeling.

Autoencoders leverage reconstruction-based objectives to learn compact latent spaces that are effective for anomaly detection, denoising, and dimensionality reduction. Their ability to uncover underlying structure from noisy or incomplete data makes them well-suited for medical signals, IoT monitoring, and RF classification tasks. Vision Transformers, originally developed for visual tasks, are increasingly adapted to sequential signal data due to their self-attention mechanisms, which allow modeling of long-range temporal dependencies and spatial correlations—especially when signals are transformed into time-frequency images such as spectrograms or scalograms.

This review aims to provide a cross-domain synthesis of how AEs, ViTs, and their hybrid forms are applied in unsupervised and semi-supervised settings for time-series signal analysis. It identifies recent methodological innovations, summarizes performance across benchmark datasets, and highlights opportunities for future research. Specifically, this review is guided by the following questions:

1) Why are Autoencoders and Vision Transformers particularly effective for time-series data?

2) Which tasks are most frequently addressed using these models?

3) What public datasets are most often used in these domains?

4) How have researchers adapted ViTs and AEs to better suit signal data?

By addressing these questions, we aim to connect theoretical developments with practical implementations across four key domains: medical signals, IoT time series, wireless signals, and radar data. The remainder of this paper is organized as follows: Section 2 introduces theoretical foundations, Section 3 surveys methodologies by domain, Section 4 reviews applications, Section 5 presents a comparative cross-domain analysis, Section 6 discusses challenges and future directions, and Section 7 concludes.

## Contribution and Novelty

This review sets itself apart by offering a comprehensive and comparative perspective on the use of AEs and ViTs in unsupervised signal classification, an area where the intersection of these two powerful architectures has received limited consolidated attention. While existing surveys have typically focused on Autoencoders in the context of general time-series modeling [1] [2] or on ViTs in computer vision [3] [4], there remains a noticeable gap in synthesizing their roles and performance across signal domains.

What distinguishes this work is its cross-disciplinary scope. By examining applications in biomedical signal processing, wireless communications, radar-based sensing, and Internet of Things (IoT) monitoring, we highlight how these models—often developed within isolated research communities—are converging around shared methodological themes. Among these are the use of masked pretraining strategies [5], hybrid ViT-AE architectures that balance reconstruction with global context modeling [6], and spectral-domain encoding techniques for time-series signals [7].

Beyond cataloging model architectures, we delve into the practical design choices that shape performance, including signal transformation methods, patch configuration, and the interplay between data sparsity and model robustness. This analysis brings to light consistent patterns and trade-offs across domains, offering insights into how best to adapt these models for real-world deployment.

To the best of our knowledge, this is the first review to bring together the rapidly evolving literatures on AEs and ViTs under a unified lens focused on unsupervised time-series analysis. Our contribution lies not only in mapping technical advancements but also in identifying future directions that can foster greater model generalization, interpretability, and cross-domain transferability in signal processing applications.

## 2. Background

AEs and ViTs have significantly transformed unsupervised learning for signal classification, outperforming traditional approaches such as K-Means and Principal Component Analysis (PCA) in modeling complex, high-dimensional data. AEs are particularly effective in tasks involving reconstruction and feature learning, with advanced variants like Variational Autoencoders (VAEs) and Masked Autoencoders (MAEs) extending their utility to generative modeling and self-supervised learning [5] [8]. On the other hand, ViTs leverage self-attention mechanisms to capture long-range dependencies, which is especially advantageous in processing spatiotemporal signals [9]. There are also studies analyzing the self-attention mechanism and positional encoding in transformer-based models [10].

Hybrid AE-ViT architectures have further advanced the field by integrating the representational power of AEs with the contextual awareness of ViTs, yielding state-of-the-art results in challenging domains such as medical diagnostics [5], radar recognition [11], and IoT monitoring [12]. This synergy forms the theoretical and practical basis for analyzing signal classification across domains in the subsequent sections.

This section outlines the theoretical underpinnings of unsupervised learning, focusing on AE and ViT-based architectures. We begin with a brief review of traditional clustering and dimensionality reduction techniques, followed by a detailed discussion of AE and ViT frameworks, including their mathematical formulations. This foundation emphasizes the suitability of these models for capturing intricate patterns in high-dimensional, time-series signal data.

### 2.1. Traditional Unsupervised Learning Techniques

Classical unsupervised learning algorithms, such as K-Means [13], Principal Component Analysis (PCA), and Density-Based Spatial Clustering of Applications with Noise (DBSCAN) [14], have been foundational for clustering and dimensionality reduction. These methods are computationally efficient but face limitations with the non-linear, high-dimensional data common in time-series, biomedical signals, and wireless communications. K-Means assumes isotropic clusters, struggling with overlapping or non-spherical distributions. PCA, effective for linear dimensionality reduction, cannot capture intricate non-linear relationships. DBSCAN excels at detecting arbitrarily shaped clusters but falters with varying density levels, a frequent challenge in heterogeneous signal data [15].

## 2.2. Autoencoder-Based Unsupervised Learning

AEs are neural networks that learn compressed, information-rich representations of input data in an unsupervised manner [1]. Unlike PCA, AEs model non-linear dependencies, making them ideal for tasks like reconstruction, anomaly detection, and feature learning in complex signals. In signal classification, AEs extract robust latent features, enhancing performance in domains like ECG analysis [16] and IoT anomaly detection [17].

Several AE variants address specific signal characteristics:

- **Convolutional Autoencoders (CAEs)** [18]: Capture local spatial patterns in image-like signal representations, such as spectrograms.
- **Long Short-Term Memory Autoencoders (LSTM-AEs)** [19]: Model temporal dependencies in sequential data, suitable for time-series signals.
- **Adversarial Autoencoders (AAEs)** [16]: Integrate generative modeling for enhanced latent space structure.
- **Variational Autoencoders (VAEs)** [8]: Use probabilistic latent spaces for structured feature learning.

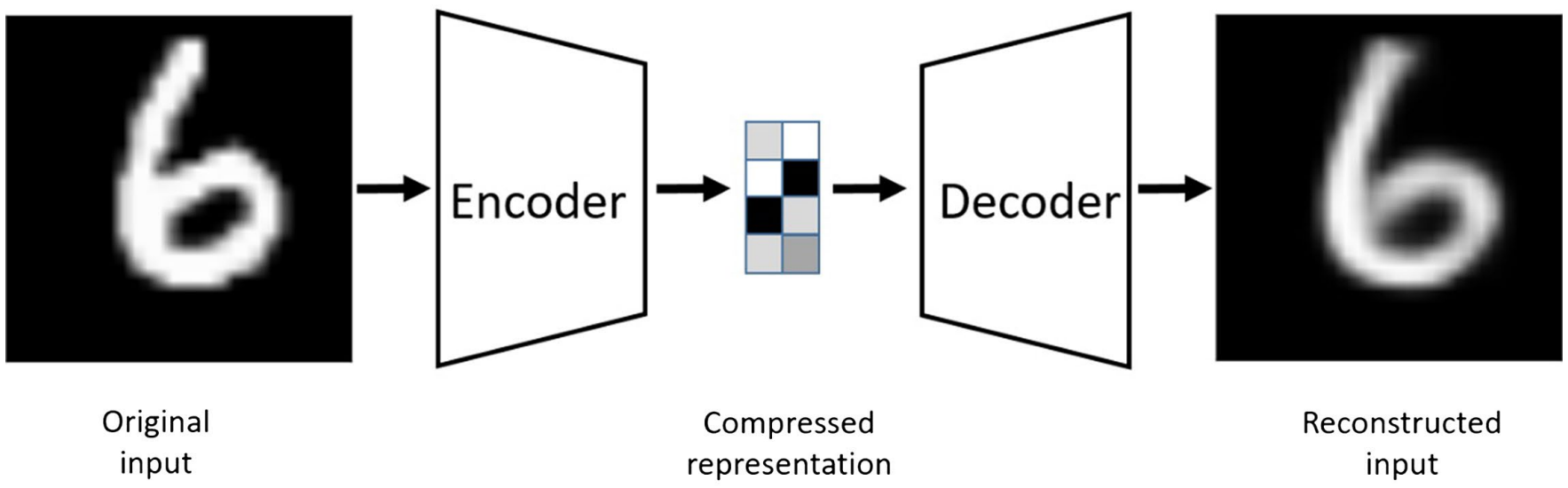

**Figure 1.** Basic Autoencoder architecture, showing encoder and decoder components [22].

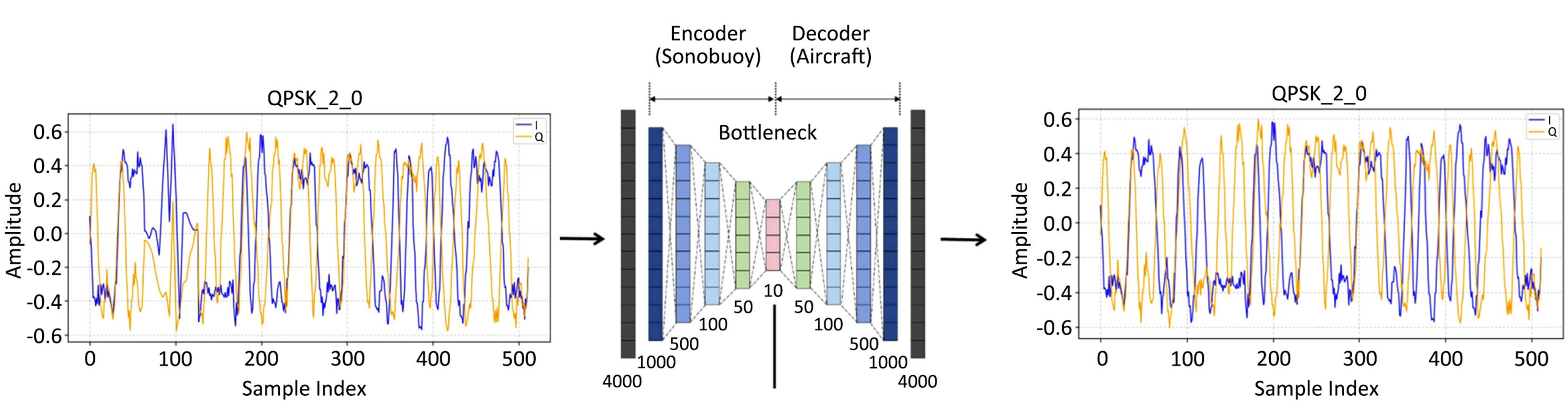

**Figure 2.** AE-based framework for signal reconstruction, highlighting latent space compression.

Recent advancements include attention mechanisms [20], low-rank attention [21], and AE-Transformer hybrids [6], improving fine-grained feature extraction. **Figure 1** and **Figure 2** illustrate AE architectures for signal processing.

### 2.3. Unsupervised Learning with Vision Transformers

ViTs, originally developed for computer vision [3], have gained prominence in signal processing due to their self-attention mechanisms, which model long-range dependencies. ViTs process inputs as tokenized patches, capturing both local and global structures in data like raw waveforms or spectrograms [7]. In unsupervised signal analysis, ViTs excel in applications such as ECG interpretation [9], IoT anomaly detection [12], and wireless modulation classification [23].

Hybrid AE-ViT models combine AE reconstruction with ViT's expressive power. Examples include NMFormer [6] for noisy modulation classification, Deno-MAE [24] for self-supervised denoising, and ViT classifiers with AE pre-training [25]. **Figure 3** shows a standard ViT adapted for signal inputs, while **Figure 4** illustrates a hybrid MAE-EEG-Transformer framework.

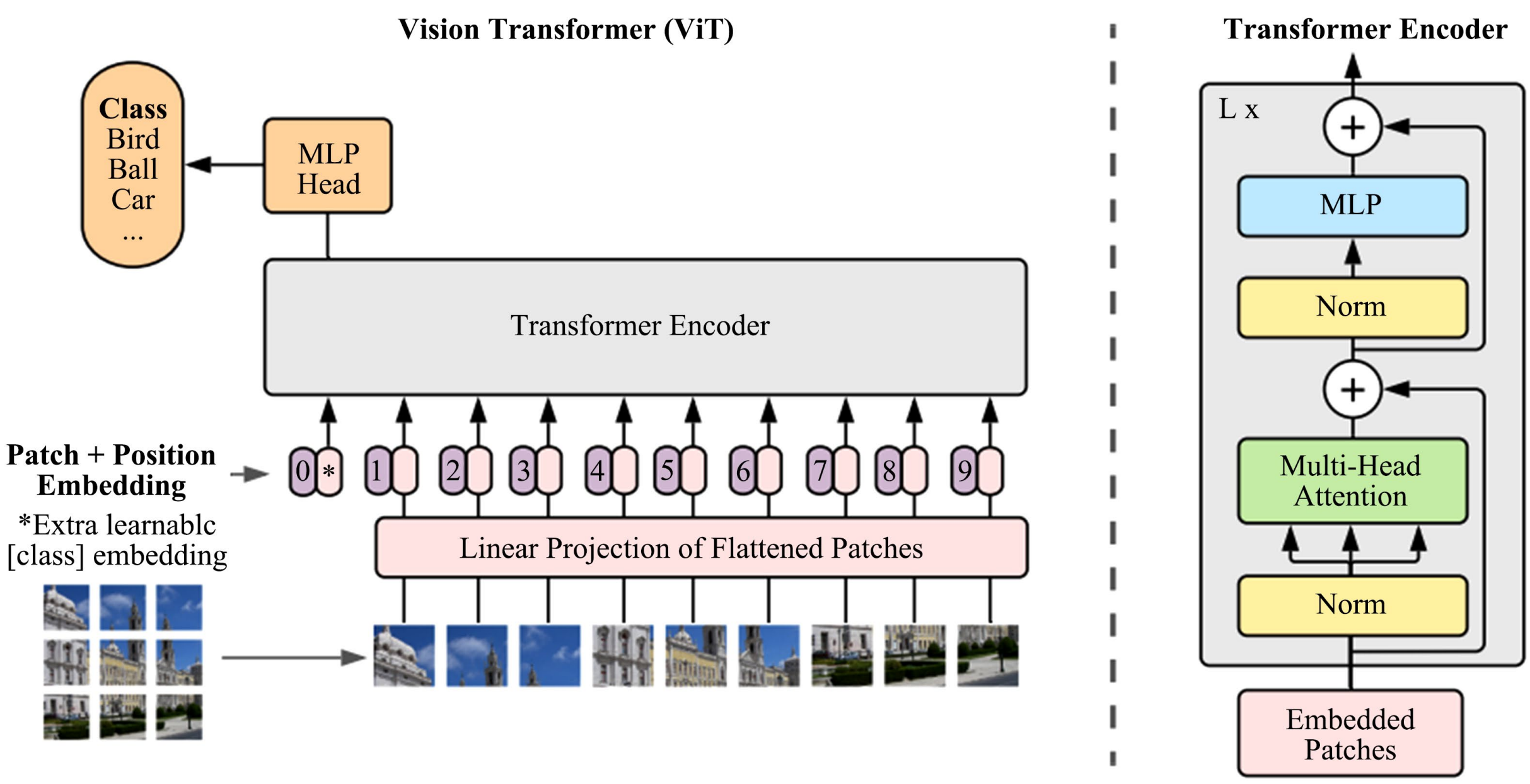

**Figure 3.** Vision Transformer architecture, processing signal inputs as tokenized patches [3].

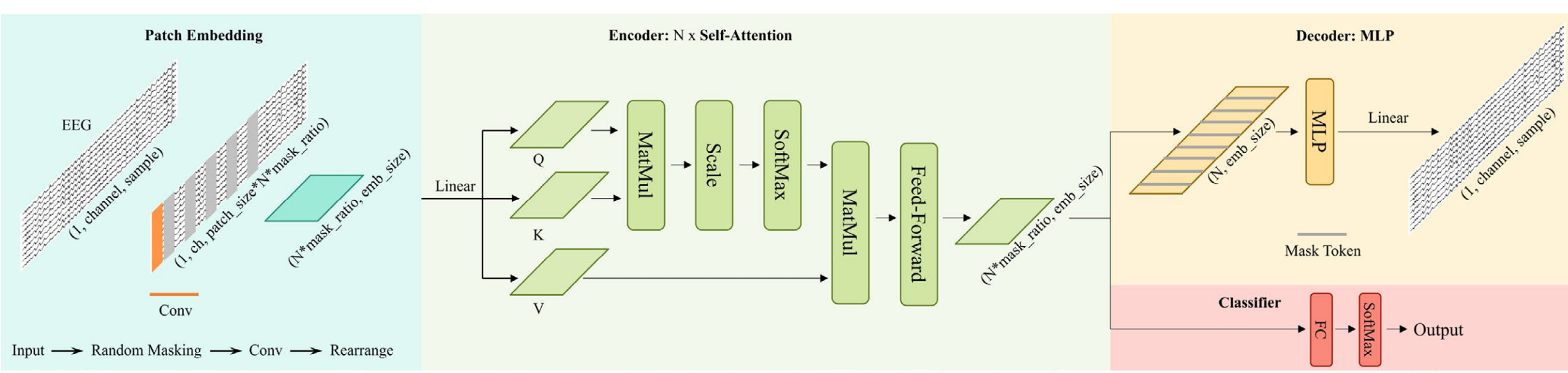

**Figure 4.** MAE-EEG-Transformer, integrating masked autoencoder pretraining for EEG classification [26].

## 2.4. Core Model Formulations

This subsection presents the mathematical foundations of AEs and ViTs, focusing on their unsupervised learning mechanisms.

### 2.4.1. Autoencoders

Autoencoders encode input $x$ into a latent representation $z = f(x;\theta_e)$ and decode it to reconstruct $\hat{x} = g(z;\theta_d)$, minimizing the reconstruction loss:

$$\mathcal{L}_{\text{AE}} = \|x - \hat{x}\|^2 \tag{1}$$

This loss ensures the reconstructed output closely matches the input, capturing essential features [1].

### 2.4.2. Variational Autoencoders (VAEs)

VAEs impose a probabilistic prior on the latent space, with the encoder outputting parameters $\mu$ and $\sigma^2$ for a Gaussian distribution:

$$z \sim \mathcal{N}(z;\mu,\sigma^2 I) \tag{2}$$

The loss, based on the Evidence Lower Bound (ELBO), balances reconstruction and regularization:

$$\mathcal{L}_{\text{VAE}} = \mathbb{E}_{q(z|x)}[\log p(x|z)] - D_{\text{KL}}(q(z|x)\,\|\,p(z)) \tag{3}$$

The ELBO encourages a structured latent space, useful for generative tasks [27]. **Figure 5** visualizes a VAE's framework.

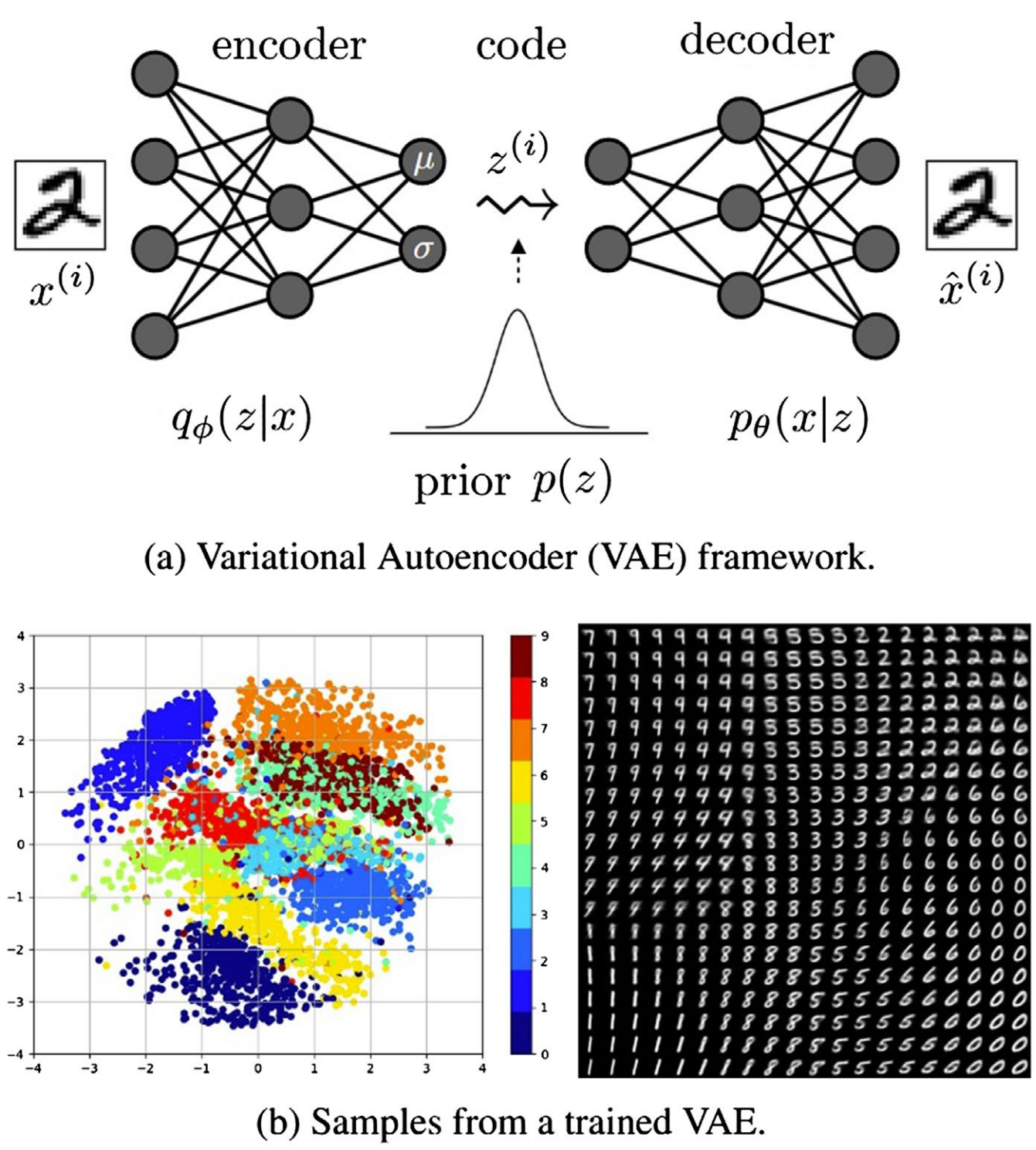

**Figure 5.** Variational Autoencoder framework, showing probabilistic latent space sampling [28].

#### 2.4.3. Adversarial Autoencoders (AAEs)

AAEs use adversarial training to align the latent space with a prior distribution, combining reconstruction and discriminator losses:

$$\min_{\theta_e,\theta_d} \mathcal{L}_{\text{AE}} \Big/ \max_{\psi} \mathbb{E}_{z\sim p(z)}\left[\log D(z)\right] + \mathbb{E}_{x\sim p_{\text{data}}(x)}\left[\log\left(1 - D\left(q(z \mid x)\right)\right)\right] \quad (4)$$

This approach enhances latent space robustness [16].

#### 2.4.4. Masked Autoencoders (MAEs)

MAEs, effective for self-supervised learning, reconstruct randomly masked input patches. The workflow includes:

1) Splitting input into fixed-length patches.
2) Masking a subset (e.g., 75%) of patches.
3) Encoding visible patches.
4) Reconstructing the full input.

MAEs excel in pretraining for signal classification [5].

#### 2.4.5. Vision Transformers

ViTs apply scaled dot-product attention to input tokens:

$$\text{Attention}(Q,K,V) = \text{softmax}\left(\frac{QK^{\text{T}}}{\sqrt{d_k}}\right)V \quad (5)$$

Multi-head attention enhances representational diversity:

$$\text{MultiHead}(Q,K,V) = \text{Concat}\left(\text{head}_1,\cdots,\text{head}_h\right)W^O \quad (6)$$

$$\text{head}_i = \text{Attention}\left(QW_i^Q, KW_i^K, VW_i^V\right) \quad (7)$$

This architecture captures rich temporal and spatial dependencies [3].

## 3. Methodologies

In this section, we provide a comprehensive overview of unsupervised learning methodologies applied to signal classification, structured into four primary categories: Medical Signals, IoT Time Series, Wireless Signals and Radar Signal Processing. For each category, we explore the application of ViTs, AEs and the combination of both (Hybrid), followed by summary tables that compare methodologies, datasets, and associated tasks.

### 3.1. Medical Signals

Analyzing medical signals such as electrocardiograms (ECG) and electroencephalograms (EEG) requires models capable of capturing both local variations and global temporal patterns. Self-attention-based architectures, particularly ViTs, have proven effective in this context by enabling richer feature extraction from complex physiological data than traditional convolutional approaches.

HeartBEiT [29] applies masked image modeling to ECG signals transformed into 2D formats, demonstrating the benefits of domain-specific pretraining. By masking parts of ECG images during training, the model enhances robustness and generalization in downstream tasks.

Liu *et al.* [30] proposed ECVT-Net, a hybrid deep learning model that integrates CNNs and ViTs for detecting Congestive Heart Failure (CHF). CNN layers extract localized ECG features, which are then divided into patches and processed by a ViT to capture long-range dependencies. This architecture achieved 98.88% accuracy in inter-patient CHF classification and remained resilient under noisy conditions.

Jamil *et al.* [31] introduced a ViT-based framework for detecting Valvular Heart Disease (VHD) using phonocardiogram (PCG) signals. The model converts raw PCG signals into time-frequency representations (TFR) using Continuous Wavelet Transform (CWT) and learns spatial representations directly from these images. It achieved 99.9% classification accuracy, outperforming CNN-based models in both performance and efficiency.

Banville *et al.* [32] employed self-supervised learning for EEG representation extraction using pretext tasks like contrastive predictive coding (CPC), relative positioning (RP), and temporal shuffling (TS). These approaches supported sleep staging and pathology detection in label-scarce environments.

Telangore *et al.* [33] presented a multimodal ViT-CNN framework for early prediction of sudden cardiac death (SCD). The model combines 1D-CNNs and LSTMs for temporal feature extraction from raw ECG, and uses ViTs with 2D-CNNs for processing scalograms and Hilbert-Huang spectra. It achieved 98.81% accuracy, demonstrating strong predictive potential for early clinical interventions. The workflow of this multimodal framework is illustrated in **Figure 6**.

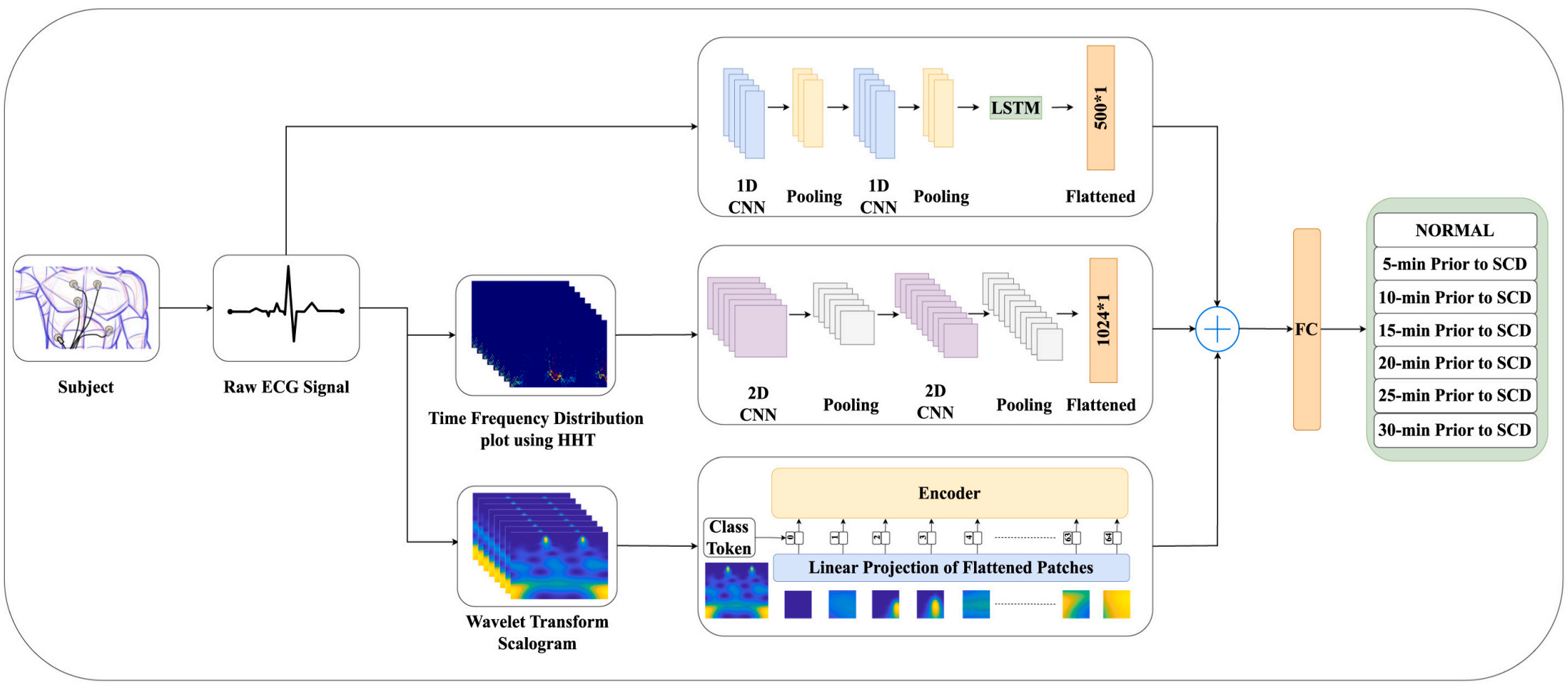

**Figure 6.** Workflow of the proposed multi-modal feature fusion and sudden cardiac death prediction framework [33].

AEs have also been widely adopted for analyzing medical signals, offering capabilities in anomaly detection, feature learning, clustering, and reconstruction. Their ability to learn nonlinear latent representations makes them well-suited for modeling noisy and complex biomedical data such as ECG and EEG.

Nejedly *et al.* [34] developed a temporal autoencoder for semi-supervised clus-tering and classification of intracranial EEG (iEEG). By compressing temporal features and applying kernel density estimation, the model supports large-scale neurophysiological data analysis with minimal labels.

Similarly, Deldadehasl *et al.* [35] introduced a dynamic classification model based on vector quantization and Lyapunov-stable gradient systems for breast cancer diagnosis. Their approach adapts over time to the input distribution and shows strong clustering performance on the high-dimensional Wisconsin Diagnostic Breast Cancer dataset, highlighting the relevance of adaptive unsupervised methods for complex biomedical signal spaces.

DeepAnT [36], initially designed for general time series, has shown strong performance on medical signals like ECG and EEG. It forecasts future signal values and flags deviations as anomalies, enabling robust unsupervised detection with a convolutional backbone.

Jang *et al.* [36] introduced a Convolutional Variational Autoencoder (CVAE) that models the variability in ECG patterns through learned latent distributions, facilitating clustering and anomaly detection without requiring annotations.

Shan *et al.* [16] used Adversarial Autoencoders (AAEs) with Temporal Convolutional Networks (TCNs) to analyze ECG signals. Their model improves anomaly detection by combining adversarial regularization with reconstruction loss and discriminator scores, helping reduce overfitting in unsupervised settings.

Zhao *et al.* [38] proposed a method for ECG signal quality assessment based on AE reconstruction loss and likelihood-based scoring. Metrics such as AE-logMSE and AE-LLH outperform conventional quality indicators across datasets including CinC, Sleep, and Stress ECG.

In EEG-based emotion recognition, [39] proposed an autoencoder-based approach that learns subject-specific frequency bands from power spectral density (PSD) instead of using fixed spectral bands. The model yielded 40% higher classification accuracy over traditional methods.

For epilepsy detection, Wen *et al.* [40] developed AE-CDNN, a hybrid combining autoencoders and CNNs. The model encodes EEG data through convolution and pooling, then reconstructs signals via deconvolution, outperforming PCA and Sparse Random Projection (SRP) on benchmark datasets.

Recent developments in self-supervised learning have shown the effectiveness of combining ViTs with autoencoders for medical signal analysis. In particular, Masked Autoencoders (MAEs) integrated with ViTs enable unsupervised learning from raw physiological data, especially in the absence of labels.

Sawano *et al.* [41] applied the MAE framework to pretrain ViT-Base, ViT-Large, and ViT-Huge architectures on unlabeled 12-lead ECG data. ECG signals were treated as $12 \times 5000$ matrices and segmented into $1 \times 250$ patches, with 75% masked during training. This approach improved performance on downstream tasks like Left Ventricular Systolic Dysfunction (LVSD) detection, outperforming models pretrained on generic image datasets.

Zhou *et al.* [5] proposed MTECG, a self-supervised ECG framework that segments

ECG time series into patches and reconstructs masked segments using a Transformer encoder-decoder. Like Sawano *et al.*, their method effectively captures both local and global features and performs well on clustering and anomaly detection.

Together, these studies reflect a growing trend toward hybrid ViT-AE architectures in medical AI. By combining ViTs' strength in modeling temporal dependencies with AEs' reconstruction capabilities, such models provide scalable and robust solutions for clinical signal analysis, particularly in data-limited scenarios. A summary of ViT and AE applications in medical signal processing is presented in **Table 1**.

**Table 1.** Medical signal processing papers using ViT and autoencoders.

| Dataset | Method | Task | Ref. |
|---|---|---|---|
| HeartBEiT Dataset | ViT | Pretraining for ECG Classification | [29] |
| ECG Dataset | ViT & CNN | Detection of Congestive Heart Failure | [30] |
| PCG Dataset | ViT | Detection of Valvular Heart Diseases | [31] |
| EEG Dataset | ViT (CPC, RP, TS) | Sleep Staging, Pathology Detection | [32] |
| ECG Dataset | ViT & CNN | Early prediction of sudden cardiac death | [33] |
| Intracranial EEG Dataset | Temporal Autoencoder | Clustering, Classification | [34] |
| Time-series signals (ECG, EEG) | Autoencoder | Anomaly Detection | [36] |
| ECG Dataset | CVAE | Clustering, Anomaly Detection | [37] |
| ECG Dataset | AAE (TCN integrated) | Anomaly Detection, Feature Extraction | [16] |
| CinC, Sleep, Stress ECG | AE-logMSE, AE-LLH | Signal Quality Assessment | [38] |
| EEG Dataset | AE-NN | Emotion Recognition | [39] |
| EEG Dataset | AE-CDNN | Epilepsy Detection | [40] |
| 12-lead ECG Dataset | ViT + Masked Autoencoder | LVSD Detection | [41] |
| MTECG Dataset | ViT + Masked Autoencoder | Anomaly Detection, Clustering | [5] |

**Critical Evaluation and Limitations.** While the reviewed methods show high classification accuracy and strong potential for unsupervised medical signal analysis, several limitations and trade-offs remain. ViT-based models, though powerful in capturing long-range dependencies, are computationally expensive and require careful pretraining strategies—especially in data-scarce domains like healthcare. The high memory and data demands of ViTs may limit their adoption in edge or wearable health monitoring devices. On the other hand, autoencoders offer efficiency and robust reconstruction abilities, but often struggle with interpretability and temporal dynamics unless enhanced with auxiliary modules like temporal convolutions or adversarial regularization. Hybrid AE-ViT models aim to balance these aspects, yet their increased architectural complexity can introduce training instability and longer convergence times. Furthermore, despite the progress in masked pretraining and contrastive learning, most models lack built-in interpretability mechanisms, posing risks in clinical settings where model decisions must be explainable. Therefore, while the hybridization trend is promising, further research is needed to improve generalization across patient cohorts, reduce training overhead, and enhance interpretability without compromising performance.

### 3.2. IoT Time Series

In IoT time series analysis, capturing long-range dependencies and extracting informative features from high-dimensional streams are critical tasks. Recent methods address these challenges by converting temporal data into image-like structures, allowing Vision Transformers (ViTs) to harness pretrained visual models for improved performance across applications such as anomaly detection, intrusion prevention, and activity recognition.

Li *et al.* [42] proposed ViTST, a method that converts each variable in a multivariate time series into a separate line graph, then arranges these plots into a grid to form a single image input for a Vision Transformer. This approach achieved strong results across irregular and regular time series, particularly in healthcare, and demonstrated robustness to missing data.

Ni *et al.* [43] provided a broad survey of imaging methods for time series, such as line plots, heatmaps, spectrograms, Gramian Angular Fields, and Recurrence Plots. Their study emphasized the benefits of applying ViTs to IoT data, highlighting improvements in classification accuracy and resilience to noise and incomplete data.

Zhang *et al.* [44] introduced TSVIT, an architecture combining 1D CNN layers for patch embedding with a Transformer encoder. This end-to-end model showed excellent performance in fault diagnosis within industrial IoT settings.

Sana *et al.* [12] developed an intrusion detection framework that integrates ViTs with traditional and deep learning methods to enhance anomaly detection in IoT networks. Tested on the NSL-KDD dataset, the ViT-based model achieved perfect accuracy, F1-score, AUC, and MCC, outperforming conventional models such as SVMs, LSTMs, and ensemble techniques. Bayesian optimization further improved performance, demonstrating the viability of ViTs in real-time, scalable intrusion prevention for IoT systems.

Tarasiou *et al.* [45] presented a Transformer model for Satellite Image Time Series (SITS) that factorizes attention both temporally and spatially, improving analysis for multivariate IoT datasets with temporal dependencies.

Liu *et al.* [15] conducted a comprehensive review of unsupervised deep learning methods for IoT time series, focusing on anomaly detection and clustering. Their work addressed challenges arising from the high dimensionality and complex spatio-temporal dynamics of IoT data, and highlighted research opportunities for future improvements. A conceptual framework of this analysis is shown in Figure 7. Table 2 provides a summary of ViT and autoencoder-based methods applied to IoT time series.

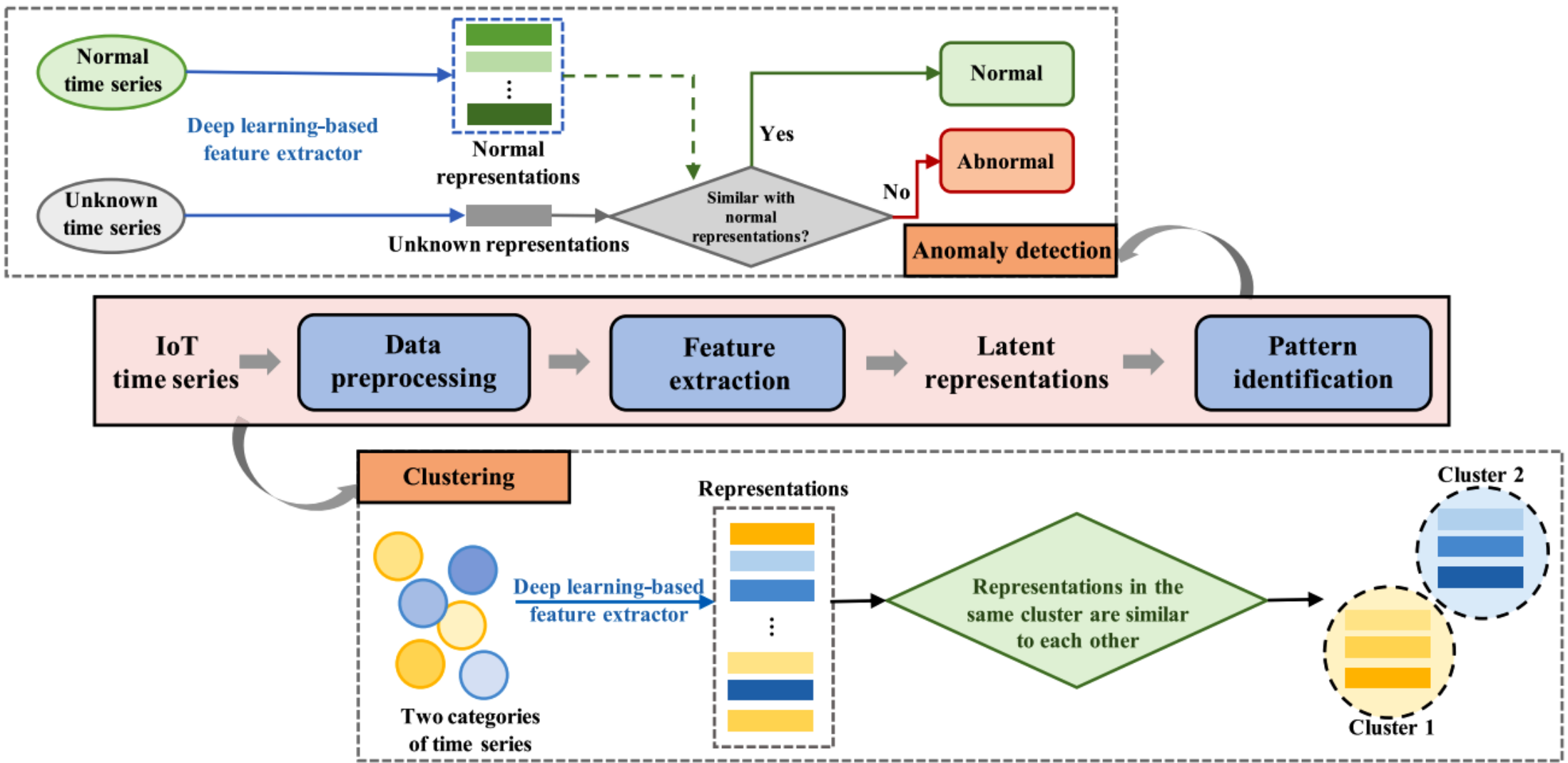

Figure 7. Example of Iot time-series analysis [15].

Table 2. Summary of ViT-based methods and autoencoder-based methods applied to IoT time series.

| Dataset Name | Reference | Method | Task |
|---|---|---|---|
| Multivariate IoT Time Series | [42] | ViTST | Time Series Classification |
| Satellite Image Time Series (SITS) | [45] | ViTSITS | Multivariate Time Series Analysis |
| Vibration Signal Dataset | [45] | TSVIT | Fault Diagnosis |
| IoT Time Series Survey | [43] | Survey | Imaging Techniques for IoT Data |
| NSL-KDD | [12] | Vision Transformer (ViT) | Intrusion Anomaly Detection |

**Continued**

| | | | |
|---|---|---|---|
| IoT Time Series Dataset | [15] | Autoencoder, GAN, CNN, RNN | Anomaly Detection, Clustering |
| Various IoT Monitoring Systems | [36] | DeepAnT | Anomaly Detection |
| MOD, ACIDS, RealWorld-HAR, PAMAP2 | [46] | FreqMAE | Vehicle Classification, Human Activity Recognition |
| Industrial Paste Thickener | [47] | Contrastive Blind Denoising Autoencoder | Real-Time Denoising |
| Cowrie Honeypot-based IoT Attack Dataset | [48] | Autoencoder | Unsupervised Clustering of IoT Attacks |
| Intel Lab Data | [49] | DTDA-RNI | Data Compression and Reconstruction |
| NSL-KDD | [50] | AAE, KNN | Intrusion Detection on IoT Edge Devices |
| Custom Dataset | [17] | VAE | IoT Traffic Anomaly Detection |
| Server Machine Dataset, Air Quality Index | [51] | R-CVAE + Transformer | Anomaly Detection |
| N-BaIoT, CICIoT2022 | [52] | VAE + ViT | IoT Botnet Detection |

AEs have also been widely applied to IoT time series for tasks such as anomaly detection, denoising, and latent feature extraction. Their ability to learn compact representations makes them effective in handling high-dimensional, noisy IoT data streams.

Liu *et al.* [15] demonstrated the use of AEs, GANs, CNNs, and RNNs for clustering and anomaly detection, focusing on the scalability of these models in heterogeneous IoT networks.

DeepAnT [36], a convolutional autoencoder that forecasts future time points and flags deviations as anomalies, has been effectively used for IoT monitoring across multiple deployments.

Kara *et al.* [46] proposed FreqMAE, a self-supervised masked autoencoder that incorporates domain-specific signal processing for multi-modal IoT environments. Its architecture includes a Temporal-Shifting Transformer (TS-T), factorized fusion mechanism, and a frequency-weighted loss, allowing robust representation learning without requiring labels.

Haseeb *et al.* [48] introduced an AE-based feature construction method to cluster IoT cyberattacks. Their model maps command features to a latent space, enabling more meaningful behavioral clustering than traditional techniques.

Langarica and Núñez [47] developed the Contrastive Blind Denoising Autoencoder (CBDAE), which applies noise contrastive estimation to regularize latent space during training, allowing real-time denoising without clean signal references in industrial IoT settings.

Xin *et al.* [49] presented DTDA-RNI, a lightweight denoising autoencoder framework for compressing and cleaning noisy sensor data. By using random noise injection (RNI) during training, the model improved reconstruction accuracy and efficiency compared to compressed sensing approaches, making it suitable for low-bandwidth IoT applications.

Aloul *et al.* [50] proposed an intrusion detection system using Adversarial Autoencoders (AAEs) combined with a KNN classifier. Deployed on a Raspberry Pi 3B, the model achieved 99.991% accuracy and operated with minimal latency, demonstrating feasibility for edge IoT devices.

Xin *et al.* [17] designed a hybrid CNN-VAE model for IoT traffic classification and anomaly detection. Reconstruction loss and KL divergence were used to detect abnormal behavior, and Particle Swarm Optimization (PSO) was applied to optimize the deep autoencoder structure.

Yao *et al.* [51] proposed THREADS, a hierarchical anomaly detection system for IIoT combining edge-deployed R-CVAEs with a cloud-based transformer discriminator. Their dual-thread architecture achieved strong performance while reducing resource usage on constrained devices.

Wasswa *et al.* [52] compared ViT and VAE encoders for IoT botnet detection. VAEs consistently outperformed ViTs on structured network data due to their strength in modeling non-visual, tabular input, highlighting VAE's advantage in cybersecurity applications over spatially biased ViTs.

**Critical Evaluation and Limitations.** Despite the promising performance of ViT-based methods in IoT time series, several practical challenges persist. ViTs typically require transformation of sequential data into image-like formats, which, while enabling reuse of visual backbones, can introduce spatial biases and obscure temporal causality. These models also demand high computational resources, making real-time inference on edge devices or constrained IoT hardware challenging. Conversely, autoencoders, especially lightweight or denoising variants, are better suited for edge deployment due to their simplicity and lower latency. However, they often struggle with learning cross-modal or long-range dependencies unless augmented with auxiliary modules like RNNs or temporal attention mechanisms. Moreover, most ViT-based models lack explicit mechanisms for uncertainty estimation or anomaly explanation, which are critical for trustworthy IoT anomaly detection in mission-critical systems. While hybrid AE-ViT models attempt to bridge these gaps, their increased complexity can hinder deployment and require careful tuning to balance generalization and resource efficiency.

### 3.3. Wireless Signals

Wireless signal processing tasks—ranging from automatic modulation classifica-tion (AMC) to anomaly detection and RF signal generation—benefit from models that can handle complex temporal and spectral patterns. Leveraging self-attention mechanisms, ViTs excel at modeling high-dimensional data, particularly when signals are represented as time-frequency diagrams or image-like formats.

Chen *et al.* [53] proposed an unsupervised RF fingerprinting framework that addresses the domain shift challenge caused by variations in channel conditions and environmental factors. Their method leverages contrastive learning and processes spectrograms of raw RF bursts using a Vision Transformer (ViT) combined with momentum contrast. The model introduces pseudo-labeling and phase-preserving augmentations to enrich the training signal without requiring ground truth labels. Unlike prior domain adaptation techniques that rely on alignment strategies, strong domain-specific assumptions, or adversarial training, this approach offers a simpler and more robust solution. The authors define positive pairs as RF signals from the same transmission and negative pairs as those from different transmissions, guiding the model to learn domain-invariant representations. Experiments on a 200-device RF testbed demonstrated an accuracy of 92.3%, significantly outperforming supervised CNNs in low-label regimes. This work represents the first application of contrastive learning for domain adaptation in RF device fingerprinting, highlighting its potential for scalable and generalizable wireless device classification.

In contrast to prior domain adaptation methods, which often depend on alignment techniques, domain-specific assumptions, or adversarial training, their method introduces a simpler and more stable alternative based on contrastive learning. This self-supervised framework uses a pretext task to bring signals from the same transmission (positive pairs) closer and push apart signals from different transmissions (negative pairs) in the learned embedding space. This design enables the model to focus on discriminative, domain-invariant features, effectively mitigating the impact of domain-specific variations. Evaluations on both wireless and wired RF datasets collected over several days showed consistent and substantial accuracy improvements (10.8% to 27.8%) over baseline models. To the best of their knowledge, this work represents the first application of contrastive learning to domain adaptation in RF device fingerprinting.

Autoencoders are also widely used in wireless signal processing for tasks such as denoising, clustering, anomaly detection, and representation learning.

Bai *et al.* [54] proposed an unsupervised autoencoder framework that uses Random Fourier Feature embeddings for clustering modulation signals. Combined with a novel separable loss function, their model achieved 15% - 20% higher clustering accuracy on the RadioML2016 dataset and remained robust across SNR conditions from 0 - 20 dB.

Chen *et al.* [55] explored self-supervised learning for RF-based human activity recognition using CSI perturbations. Their method extracts spatio-temporal features from raw RF data without requiring sensors or labels, achieving strong performance in gesture recognition and occupancy detection.

Faysal *et al.* [24] introduced DenoMAE, a multimodal framework that integrates a ViT branch for time-frequency input and a convolutional AE branch for raw waveforms. Using masked self-supervised learning, it achieved 15% - 20% higher classification accuracy and 3 - 5 dB PSNR noise reduction under low-SNR conditions. The pretraining strategy of DenoMAE is shown in **Figure 8**.

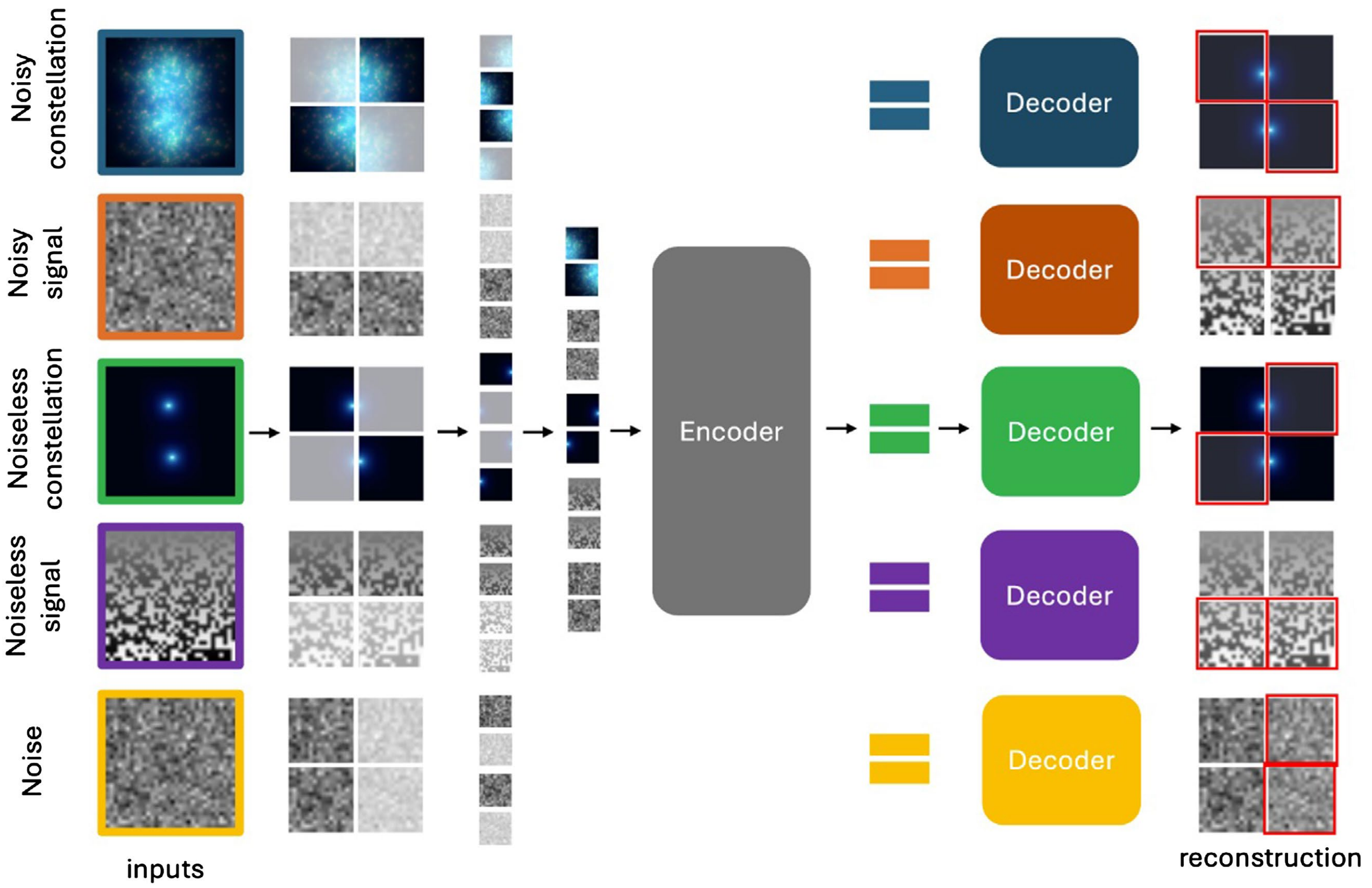

**Figure 8.** DenoMAE Pretraining Strategy: A random 75% masking is applied across all input modalities (not to scale). The remaining 25% of visible patches are passed through a shared encoder. Each modality then uses its own decoder to reconstruct the masked portions. Only the encoder is reused during downstream fine-tuning [24].

Shi *et al.* [56] proposed GAF-MAE, which transforms time-series RF signals into Gramian Angular Fields (GAF) and trains a ViT-based masked autoencoder for robust reconstruction. The model improved AMC accuracy by 12% - 18% over CNNs in 5 - 15 dB SNR conditions.

Gupta *et al.* [57] presented SpectraViT, a hybrid ViT-ConvAE framework for unsupervised spectrum anomaly detection. Operating across 1 MHz to 6 GHz, it achieved 94% AUC in detecting signal interference in real-time.

Lee and Park [58] developed a self-supervised ViT framework using masked autoencoding on spectrogram patches. It outperformed supervised CNNs in modulation recognition, achieving 85% accuracy across 24 modulation types even with limited labeled data.

Collectively, these studies (summarized in **Table 3**) demonstrate growing interest in leveraging ViTs and AEs for wireless signal analysis. Applications span from AMC and anomaly detection to RF fingerprinting and spectrum monitoring. ViTs offer strong modeling of global features, while AEs provide powerful reconstruction and denoising capabilities. Their integration—often through masked modeling, hybrid architectures, or multimodal fusion—offers a robust and scalable approach to wireless communication systems.

**Table 3.** Wireless signal processing papers using ViT and autoencoders.

| Dataset | Method | Task | Ref. |
|---|---|---|---|
| RML2016.10a | GAF-MAE (ViT + AE) | AMC | [56] |
| RadioML2018 | DenoMAE (ViT + AE) | Denoising | [24] |
| RadioML2016 | RffAe-S (AE) | Clustering | [54] |
| Custom RF | ViT + Contrastive Learning | RF Fingerprinting | [53] |
| Spectrum Monitoring | SpectraViT (ViT-ConvAE) | Anomaly Detection | [57] |
| Simulated | ViT + Masked Autoencoding | Modulation Recognition | [58] |
| Wi-Fi/mmWave CSI | AE + Self-Supervised | Human Activity Recognition | [55] |

**Critical Evaluation and Limitations.** While ViT- and AE-based models have shown impressive results in wireless signal processing, their practical deployment still faces notable trade-offs. Vision Transformers, especially in contrastive and masked self-supervised settings, offer powerful global feature modeling but often rely on spectrogram or image transformations that may obscure phase information or temporal locality critical in RF tasks like modulation recognition. Additionally, ViTs are typically resource-intensive and less suited to real-time or embedded spectrum monitoring systems. Autoencoders, particularly when adapted for spectral or time-series inputs, are more lightweight and efficient but often lack robustness under high-noise or low-SNR conditions unless paired with advanced regularization techniques. Moreover, while hybrid ViT-AE models offer a compelling compromise, their multimodal fusion schemes and decoder complexity can increase the risk of overfitting and require careful hyperparameter tuning. Importantly, most current models are trained and evaluated under static or lab-controlled conditions; thus, their generalizability to dynamic wireless environments with mobility, interference, and protocol variation remains an open challenge.

### 3.4. Radar Signals

Radar signal processing presents unique challenges due to low signal-to-noise ra-tios, environmental complexity, and limited labeled data. Recent advances in unsupervised learning—especially ViTs, AEs, and their hybrids—have enabled progress in tasks such as clutter suppression, target detection, waveform recognition, and human activity monitoring across diverse radar modalities including SAR, PolSAR, FMCW, IR-UWB, and HRRP.

ViTs leverage global self-attention mechanisms, enabling effective modeling of long-range dependencies and outperforming CNNs in many radar scenarios.

Kim *et al.* [59] developed a ViT-based approach for waveform recognition of Low Probability of Intercept (LPI) radar signals, achieving a 12.8% accuracy improvement at −10 dB SNR. Kayacan and Erer [60] proposed Declutter ViTs (DC-ViTs) for clutter removal in Ground Penetrating Radar (GPR), improving signal-to-clutter ratio by 20%.

Ghosh and Bovolo [61] introduced a self-supervised ViT with contrastive learning for radargram segmentation, increasing MIoU by 23.47%. Yu *et al.* [62] presented SLViT, a multimodal slot-based ViT architecture that mitigates speckle noise in SAR classification through modality-specific slots.

Li *et al.* [63] proposed PolSAR-MPIformer, leveraging mixed patch interactions for fusing dual-frequency PolSAR images. Wang *et al.* [64] combined Swin Transformers and supervised contrastive learning in SCL-SwinT for robust human activity recognition from IR-UWB radar.

Muzeau *et al.* [65] introduced SAFE, a masked Siamese ViT framework with SAR-specific augmentations, achieving strong generalization across unseen sensors. Li *et al.* [66] proposed MTBC, which integrates a masked ViT with Brown distance covariance for few-shot recognition of HRRP radar data.

Feng *et al.* [67] combined ViTs with Canny edge detection for unsupervised SAR interference pattern analysis. Shi *et al.* [68] proposed a semi-supervised ViT model for FMCW radar hand gesture recognition using pseudo-label consistency to improve performance with limited labeled data.

Complementing ViTs, AEs are widely used in radar signal processing for their strengths in denoising, anomaly detection, and latent representation learning in cluttered environments.

Xiang *et al.* [69] developed MVN (Masked Phase Autoencoder with ViT) for radar-based heart rate estimation, enhancing accuracy while reducing observation time. Dal Corso and Bruzzone [70] used autoencoders trained with random walks for unsupervised segmentation of radar sounder data, achieving efficient label propagation.

Kahya *et al.* [71] proposed a lightweight AE model for out-of-distribution detection in short-range FMCW radar, achieving an AUROC of 90.72%. Jiang *et al.* [72] combined AEs with LSTMs for radar pulse pre-sorting, improving classification of low-frequency pulses.

Liang *et al.* [73] introduced GM-CVAE, a Gaussian Mixture Convolutional VAE for radar target detection in clutter, outperforming parametric methods in complex scenes.

Hybrid ViT-AE models integrate the global modeling of transformers with AE reconstruction capabilities, offering significant gains in tasks involving fusion, few-shot learning, and semantic understanding.

Wang *et al.* [74] proposed FG-MAEs (Feature-Guided Masked Autoencoders) for semantic SAR classification, improving accuracy by 5% over standard MAEs. Chan-To-Hing and Veeravalli [75] introduced FUS-MAE, a cross-attention MAE model for SAR-optical fusion, achieving superior cross-modal performance.

Wang *et al.* [76] designed MAPM, integrating positional prediction and memory tokens into MAEs for PolSAR classification in low-label settings. Huan *et al.* [11] built a lightweight ViT-CNN hybrid for radar-based activity recognition, balancing spatial-temporal modeling with efficiency.

These advancements, summarized in **Table 4**, illustrate the growing impact of unsupervised ViT, AE, and hybrid architectures in enabling scalable, accurate, and real-time radar signal processing.

**Table 4.** Summary of vision transformer, autoencoder, and hybrid methods for radar signal processing.

| Dataset | Method | Task | Ref. |
|---|---|---|---|
| LPI Radar | ViT | Waveform Recognition | [59] |
| GPR | DC-ViT | Clutter Removal | [60] |
| SAR | Multimodal Slot ViT | SAR Image Classification | [62] |
| PolSAR | PolSAR-MPIformer (ViT) | Adaptive Fusion Classification | [63] |
| IR-UWB Radar | SCL-SwinT (Swin Transformer) | Human Activity Recognition | [65] |
| SAR | SAFE (Masked Siamese ViT) | Self-Supervised Feature Extraction | [65] |
| HRRP | MTBC (Masked ViT + Brown Distance Covariance) | Few-shot Recognition | [66] |
| SAR Complex Imagery | Self-Supervised ViT | Interference Detection | [67] |
| FMCW Radar | Semi-Supervised ViT | Hand Gesture Recognition | [68] |
| Radar-based Heart Rate | MVN (Masked AE + ViT) | Heart Rate Estimation | [69] |
| Radar Sounder | AE + Random Walks | Semantic Segmentation | [70] |
| FMCW Radar | AE-based OOD Detector | Out-of-Distribution Detection | [71] |
| Radar Pulse | AE + LSTM | Pre-sorting | [72] |

**Continued**

| | | | |
|---|---|---|---|
| Radar Clutter | GM-CVAE (CNN + AE) | Target Detection | [73] |
| SAR | FUS-MAE (Cross-Attention MAE) | Multimodal Fusion | [75] |
| PolSAR | MAPM (Masked AE) | Image Classification | [76] |
| SAR | Feature Guided MAE | Self-Supervised Learning | [74] |
| Radar | Hybrid ViT-CNN | Activity Recognition | [11] |
| Radargram | URS (Self-Supervised ViT) | Segmentation | [61] |

**Critical Evaluation and Limitations.** Radar signal processing brings domain-specific challenges such as extreme noise, limited annotation, and high variability across modalities (e.g., SAR vs. IR-UWB). While ViT-based models have proven effective in modeling global dependencies and handling multimodal inputs, their reliance on large-scale training and complex attention mechanisms can hinder real-time applicability in embedded radar systems. Moreover, masking strategies in ViTs may lead to information loss when applied to already sparse radar returns. Autoencoders, particularly those designed for denoising or latent representation learning, provide computational efficiency and are better suited for anomaly detection in low-SNR settings. However, they often fall short in capturing spatial coherence in radar images or complex motion dynamics without architectural enhancements. Hybrid ViT-AE architectures offer balanced solutions but tend to be highly customized for specific radar types or tasks, which limits their generalizability. A key limitation across many models is the lack of robust cross-device or cross-domain validation, which is essential for deploying radar AI systems in diverse operational environments. Future research should prioritize lightweight architectures with domain adaptation capabilities and interpretable outputs, especially for safety-critical applications like autonomous sensing or disaster monitoring.

## 4. Applications

This section explores the unsupervised and semi-supervised applications of AEs, ViTs, and their hybrid approaches across four critical signal domains: Wireless Signals, Medical Signals, IoT Data, and Radar Signals. While **Figure 9** presents global trends, the focus here is specifically on tasks utilizing unsupervised or semi-supervised learning within these domains.

AEs have been extensively applied in unsupervised tasks due to their ability to learn compact and informative latent representations. They are notably effective in anomaly detection (20%), natural language and medical applications (15% each), denoising and restoration (10%), and tasks like image classification, segmentation, and audio processing (each around 10%). Their adaptability across structured and unstructured data underlines their foundational role in representation learning without supervision.

In contrast, ViTs are predominantly used in computer vision tasks, with significant application in image classification (40%) and object detection (40%). Though their use in denoising (7%) and anomaly detection (3%) is relatively limited, ViTs are increasingly favored in tasks that benefit from global spatial attention and long-range sequence modeling, expanding their relevance in complex signal interpretation.

Figure 10 illustrates the distribution of Autoencoders and Vision Transformers across supervised, unsupervised, and semi-supervised paradigms. Autoencoders dominate unsupervised use cases, accounting for about 82.5% of their applications due to their strength in learning from unlabeled data. Vision Transformers, in contrast, are mainly utilized in supervised settings (90%), reflecting their success in annotated visual benchmarks. AEs see moderate adoption in semi-supervised contexts, whereas ViTs are only recently gaining traction in such scenarios. This contrast highlights their complementary strengths: AEs in label-scarce environments, and ViTs in data-rich ones.

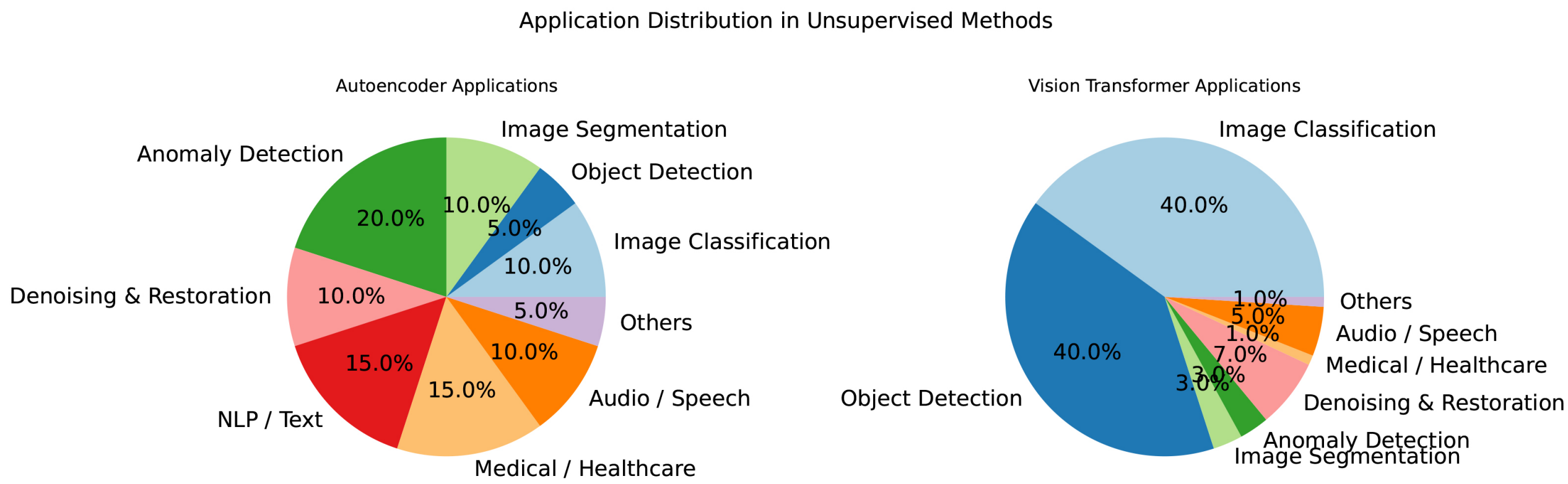

Figure 9. Applications of vision transformers and autoencoders across different domains such as wireless signals, Iot systems, radar data, and medical signals.

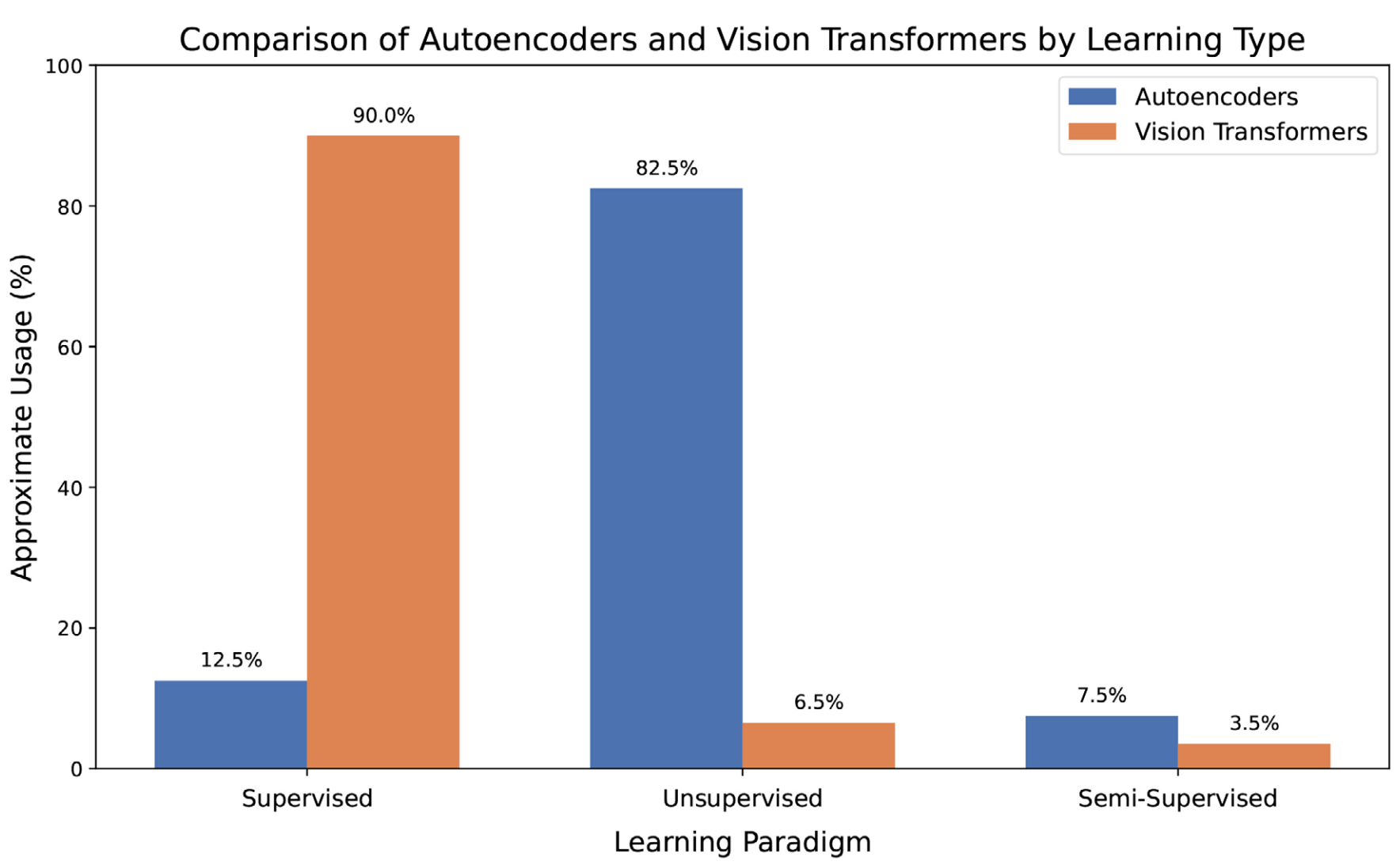

Figure 10. Comparison of autoencoders and vision transformers across learning paradigms.

Recent surveys support these trends, offering broader perspectives on AEs and ViTs. AE-focused reviews outline structural variants, evolution, and domain-spanning applications in vision, NLP, recommender systems, and anomaly detection, while identifying challenges and future directions [1] [2]. ViT surveys highlight their expanding role in digital health, visual benchmarks, and real-time applications, with focus areas including segmentation, classification, multiscale vision, and video analysis [3] [4].

## 4.1. Wireless Signal Processing

Unsupervised and self-supervised learning using AEs and ViTs have advanced wireless signal processing tasks, including modulation recognition, denoising, anomaly detection, and channel estimation. Ali and Fan [77] leveraged deep AEs for automatic modulation classification (AMC) using layer-wise pretraining across various SNR conditions.

Hybrid architectures, such as those from Shevitski *et al.* [78], combined convolutional AEs with transformers for efficient RF signal classification and demodulation. Faysal *et al.* [24] introduced DenoMAE, a masked autoencoder reconstructing clean waveforms and constellation diagrams in noisy environments.

Transformer-CNN hybrids, proposed by Qu *et al.* [79] and Zhang *et al.* [25], merged temporal self-attention with LSTM or CNNs to enhance AMC performance. Transformer-masked autoencoders (TMAEs) [80] also addressed source coding and channel estimation. Additionally, Li *et al.* [81] proposed a noise-adaptive ViT for robust AMC under adversarial attacks, and ViT-MAE models [82] outperformed CNNs in constellation-based tasks.

## 4.2. Medical Signal Processing

Medical signal analysis has seen widespread adoption of unsupervised deep learning, particularly for EEG and ECG tasks. Banville *et al.* [32] utilized contrastive and predictive self-supervised tasks on EEG, extracting latent features without labels.

Zhou *et al.* [5] introduced MTECG, a masked transformer model for ECG classification. Similarly, Sawano *et al.* [41] used ViTs with high masking ratios for ECG representation learning. Xia *et al.* [83] developed a hybrid model integrating Transformers, CNNs, and denoising AEs for arrhythmia classification, and Huan *et al.* [11] adapted ViTs for radar-based clinical activity monitoring.

## 4.3. IoT Signal Processing

In IoT contexts, where data is typically unlabeled, AEs and ViTs support anomaly detection, dimensionality reduction, and traffic classification. AE and Transformer models identify abnormal patterns without ground truth [51], while VAEs enhance botnet detection accuracy [52].

Blind denoising and contrastive AEs [47] recover multivariate sensor data robustly in noisy environments. CNN-VAE hybrids [17] detect traffic anomalies by combining supervised classification with unsupervised outlier detection via re-construction loss.

### 4.4. Radar Signal Processing

Radar processing has benefited from unsupervised techniques for clutter suppression, waveform classification, and activity recognition. Liang *et al.* [73] used a GM-CVAE to detect targets via reconstruction probabilities. Jiang *et al.* [72] proposed a CLDE (CNN-LSTM-decoder) model for radar pulse sorting in low-frequency conditions.

Kim *et al.* [59] applied ViTs to classify LPI radar using time-frequency representations, achieving robustness over CNNs. Huan *et al.* [11] developed LH-ViT, a lightweight transformer with convolutional pyramids for real-time radar-based activity recognition.

These studies underscore the versatility of AEs and ViTs in enabling adaptive, efficient signal analysis under limited supervision across diverse domains.

## 5. Comparative Study of Unsupervised Learning Across Signal Domains

This section presents a comparative synthesis of the methodologies reviewed in four signal domains: wireless communications, radar systems, IoT time series, and biomedical signals—where AEs, ViTs, and hybrid AE-ViT architectures have been applied for unsupervised or semi-supervised learning. Rather than relying on a shared benchmark, studies span a wide range of datasets—from standard repositories like RadioML, MIT-BIH, and NSL-KDD to specialized, domain-specific collections—reflecting the diversity of experimental setups and application goals.

From the tabulated summaries in the methodology section, several thematic patterns and technical trends are evident:

**Benchmark Preferences**: Certain datasets have emerged as reference points within their respective fields. For instance, RadioML is the prevailing choice for wireless modulation classification, while NSL-KDD and N-BaIoT are frequently used in IoT security applications. In medical signal processing, ECG analysis is often evaluated on MIT-BIH and CinC datasets. These standardized datasets facilitate performance benchmarking and have also played a role in shaping model pretraining practices.

**Application Breadth**: The scope of tasks tackled using AEs and ViTs is extensive. Classical objectives such as modulation classification, anomaly detection, clustering, and signal denoising appear frequently. However, more recent studies have extended into nuanced areas such as RF fingerprinting, emotion recognition, sleep stage classification, radar segmentation, and multimodal sensor fusion—demonstrating the adaptability of these models to a variety of structured and unstructured signal formats.

**Architectural Diversity**: Autoencoder-based approaches span a wide range—from basic feedforward variants to more complex designs including convolutional VAEs, adversarial AEs, and blind denoising frameworks. ViTs, while relatively newer to the signal processing domain, have already evolved to include masked autoencoding variants, hierarchical attention modules, and transformer-GAN hybrids. Notably, hybrid AE-ViT architectures are increasingly adopted to balance reconstructive learning with global feature extraction, offering promising performance in both low-data and noisy settings.

**Transformer Momentum**: Across domains, the use of ViTs is growing steadily, particularly in tasks that benefit from modeling long-range dependencies. In radar and IoT signal processing, their adoption is accelerating due to improved tokenization methods, domain-adapted augmentations, and task-specific pretraining strategies. While computationally more demanding than AEs, ViTs have demonstrated strong performance gains when signals are converted into image-like representations such as spectrograms, scalograms, or GAFs.

Despite these advances, the lack of unified cross-domain benchmarks limits broader generalization studies. This remains a key area for improvement as the field matures.

## 5.1. Representative Experimental Results

- **Medical Signals:** HeartBEiT [29] applies ViTs to masked ECG images ($16 \times 16$ patches), achieving 94.5% arrhythmia detection. ECVT-Net [30] combines CNNs with ViTs for CHF classification (98.88%), with $1 \times 250$ patch inputs. Jamil *et al.* [31] process PCG spectrograms into $32 \times 32$ patches, achieving 99.9% VHD detection. Banville *et al.* [32] use CPC and temporal shuffling for EEG (92.5% sleep staging), while Telangore *et al.* [33] integrate CNN-ViT-LSTM modules for early SCD prediction (98.81%).
- **IoT Time Series:** ViTST [42] uses $16 \times 16$ grid-encoded multivariate plots (91% accuracy). TSVIT [44] tokenizes vibration signals into $1 \times 100$ patches (93% accuracy). Sana *et al.* [12] achieve 100% metrics on NSL-KDD using ViTs with $64 \times 64$ tokenization. Tarasiou *et al.* [45] report a 5% improvement over CNNs using ViTs for satellite time series. Liu *et al.* [15] summarize 90% accuracy across various AE-based unsupervised tasks.
- **Wireless Signals:** NMformer [6] tokenizes constellation images ($16 \times 16$) for AMC with 90% accuracy. Li *et al.* [81] use $32 \times 32$ spectrogram patches for ViT-based AMR, reporting 94.17% accuracy under benign conditions and 71.2% under attack. CTGNet [84] fuses graph CNNs with ViTs (91% accuracy). RF-ViT-GAN [85] synthesizes 5G/WiFi signals with 95% fidelity. Chen and Wang [53] achieve 92.3% accuracy for RF fingerprinting using self-supervised ViTs.
- **Radar Signals:** URS [61] processes radargrams via ViT segmentation (23.4% mIoU). Kim *et al.* [86] achieve 0.96 AUC on synthetic radar data using multi-resolution ViTs. Shi *et al.* [56] utilize GAF-MAE with ViTs for AMC, showing a 15% accuracy gain over baselines. Kim *et al.* [59] apply ViTs to spectrograms of LPI radar signals (12.8% improvement over CNNs). MHSA-ViT [87] achieves 90% accuracy, though domain shift remains a challenge. See **Table 5** for representative results across different signal domains.

**Table 5.** Summary of representative experimental results across signal domains.

| Domain | Model | Dataset | Input Format | Accuracy (%) |
|---|---|---|---|---|
| Medical (ECG) | HeartBEiT [29] | MIT-BIH | 16 × 16 patches | 94.5 |
| IoT (Network) | Sana *et al.* [12] | NSL-KDD | 64 × 64 tokenization | 100.0 |
| Wireless (AMC) | NMformer [6] | RadioML | Constellation images | 90.0 |
| Radar (SAR) | MHSA-ViT [87] | Synthetic SAR | Spectrogram patches | 90.0 |

### 5.2. Key Takeaways

- **ViTs outperform in structured or visualized domains** such as medical (ECG/EEG spectrograms), radar (SAR, HRRP), and wireless (constellations, spectrograms), where signal data can be transformed into rich 2D representations. However, they underperform or require adaptation when applied to irregular tabular formats such as raw IoT logs.
- **Autoencoders remain dominant in low-SNR or embedded settings**, such as IoT anomaly detection and FMCW radar filtering, due to their lower computational burden and strong performance under resource constraints. Yet, their generalization ability and capacity to capture long-range dependencies are often limited unless enhanced with temporal modules or hybridization.
- **Hybrid AE-ViT architectures bridge the gap** by enabling both strong feature learning (ViT) and reconstruction/anomaly detection (AE), but they often require custom tuning per domain. Their benefit is most evident in multimodal or few-shot tasks such as PolSAR fusion or rare event detection in radar and IoT domains.
- **Interpretability remains an unresolved challenge**, especially in ViT-heavy models. While attention maps can offer partial insight, domain experts in healthcare or security often require more transparent and explainable decision pathways, which current models only partially address.
- **The lack of cross-domain benchmarking limits generalizability**. Most models are evaluated on single-domain datasets with little transfer learning or multi-domain validation. Establishing unified benchmarks or shared pretraining corpora could accelerate progress in generalizable signal understanding.

## 6. Challenges and Future Directions

Despite remarkable progress in unsupervised learning for signal processing—driven by AEs, ViTs, and hybrid architectures—the field continues to face technical, methodological, and domain-specific limitations. This section outlines key challenges and proposes promising research directions, including the development of signal-specific foundation models, lightweight inference strategies, and advances in scalable self-supervised learning.

Unsupervised learning in IoT environments remains constrained by the high dimensionality of time-series data, limited edge computing resources, and scarcity of labeled anomalies [51]. Robust deployment in such settings demands innovations in hierarchical architectures, adaptive latent space modeling, and noise-invariant learning strategies. Hybrid architectures that balance efficiency with expressiveness are particularly promising for real-time, resource-constrained applications [52].

Autoencoders—especially VAEs—have proven effective in dimensionality reduction, anomaly detection, and clustering, particularly in low-data or high-noise conditions. However, they often impose higher computational costs compared to classical techniques like Principal Component Analysis (PCA). Their performance is also highly sensitive to input representation, such as raw waveforms, engineered features, or time-frequency transformations—posing a barrier to generalization across heterogeneous datasets [88]. While ViTs excel at modeling long-range dependencies, they often require extensive pretraining to generalize effectively. Transfer learning from natural image datasets (e.g., ImageNet) may not yield optimal performance in domains such as ECG or RF signals. Recent studies on domain-specific pretraining—such as Masked Autoencoders tailored to physiological signals—have improved task-specific accuracy [41], but at the expense of high computational overhead and data requirements, limiting accessibility in low-resource environments.

A common workaround involves converting time-series signals into 2D visual formats (e.g., spectrograms, Gramian Angular Fields (GAF), Recurrence Plots (RP)) to enable ViT-based modeling. However, the selection of these transformations is often heuristic, lacking theoretical justification. Performance can vary significantly across transformation types, and many are non-invertible, complicating downstream tasks such as signal reconstruction, denoising, or generation [43].

In addition to these modeling challenges, unsupervised frameworks face instability in training, sensitivity to initialization, and the inherent difficulty of evaluating model quality in the absence of labels. Among these, interpretability remains a particularly pressing concern in safety-critical applications such as healthcare diagnostics, military radar, and cybersecurity.

ViT-based models often lack intuitive explanations, as attention weights may not clearly reflect decision pathways—especially when applied to transformed or noisy signals. Autoencoders, though more transparent due to their reliance on reconstruction loss, still offer limited insight into which latent features influence specific predictions. Recent efforts in explainable AI (XAI) suggest promising directions, including attention rollout, gradient-based saliency, and latent space interpolation. However, such techniques are rarely adapted or validated for signal data. Future work should focus on domain-specific interpretability methods that enable practitioners to trace model behavior back to signal components—such as frequency bands in EEG, temporal bursts in radar, or modulation patterns in RF—thereby enhancing trust, debugging, and regulatory acceptance.

### Future Research Directions

Several avenues hold strong potential for addressing the above limitations:

- **Cross-Domain Generalization:** Developing architectures that can transfer knowledge across domains (e.g., synthetic-to-real radar, ECG-to-EEG) is essential for generalizable deployment. Meta-learning and domain adaptation techniques will play a crucial role in enabling robust cross-task performance.
- **Signal-Specific Foundation Models:** Inspired by the success of foundation models in vision and NLP, pretraining large-scale, domain-tailored backbones (e.g., ViT-MAE for biosignals or RF waveforms) offers a pathway toward universal encoders for signal intelligence tasks.
- **Lightweight and Edge-Compatible Models:** The design of compact architectures suitable for deployment on embedded or IoT devices is critical. Approaches such as quantization, pruning, knowledge distillation, and transformer-efficient variants (e.g., Linformer, MobileViT) may help minimize memory and latency overheads.
- **Hybrid and Modular Architectures:** Combining CNNs for local pattern recognition with transformers for global context modeling offers a scalable strategy for diverse signal types. Modular design can support plug-and-play adaptation across tasks without requiring full retraining.
- **Scalable Self-Supervised Learning:** Advancing pretext tasks (e.g., masked token modeling, temporal contrastive learning) and augmentations (e.g., frequency masking, jittering) will improve feature quality in unlabeled settings. Efficient training strategies will be essential to extend these methods to high-volume signal streams.

While unsupervised and self-supervised frameworks have already demonstrated impressive potential across signal domains, overcoming existing challenges will require a synergy of architectural innovation, data-efficient learning, and principled evaluation. The future of signal AI lies in scalable, interpretable, and transferable models that bridge the gap between real-world constraints and high-performance learning paradigms.

## 7. Conclusions

This review has explored how unsupervised and semi-supervised learning methods—specifically Autoencoders, Vision Transformers, and their hybrid configurations—are being applied to time-series signal classification across four principal domains: wireless communications, radar systems, IoT time series, and biomedical signals. Drawing from a broad set of recent studies, we analyzed the architectural trends, domain-specific adaptations, and evolving research directions underpinning these approaches.

**1) Why are Autoencoders and Vision Transformers particularly effective for**

**time-series data?** Autoencoders have proven to be strong candidates for capturing intrinsic features in time-series signals, especially when labeled data is scarce. Their ability to compress high-dimensional inputs into informative latent spaces makes them well-suited for anomaly detection, reconstruction, and clustering. Vision Transformers, on the other hand, bring a novel perspective to time-series modeling by enabling the capture of long-range dependencies through self-attention. When applied to time-frequency or image-like signal representations, ViTs offer a powerful means to analyze global patterns that may not be easily accessible through traditional models. The combination of both models allows for enhanced flexibility in representing both local and global signal structures.

**2) Which tasks are most frequently addressed using these models?** Across the surveyed domains, these models have been widely adopted for tasks such as automatic modulation classification, anomaly detection, signal reconstruction, and noise suppression. In biomedical settings, they are frequently used for ECG and EEG analysis, including arrhythmia detection and emotion recognition. In IoT and network environments, they have found strong utility in intrusion detection and traffic analysis. Meanwhile, radar systems are increasingly leveraging these techniques for tasks such as target recognition, waveform segmentation, and scene understanding.

**3) What public datasets are most often used in these domains?** A number of benchmark datasets have become central to model evaluation. These include RadioML for modulation classification, MIT-BIH for ECG analysis, NSL-KDD and N-BaIoT for IoT security, and synthetic radar datasets for waveform analysis. While these datasets have helped standardize evaluation, there remains a strong need for more diverse, high-resolution datasets that better reflect real-world variability and conditions.

**4) How have researchers adapted ViTs and AEs to better suit signal data?** Recent innovations include the use of denoising and variational autoencoders, adversarial training techniques, and CNN-AE hybrids for learning richer temporal features. For ViTs, adaptations have involved the use of patch-based encodings of spectrograms and time-series images, masked pretraining strategies, and temporal or frequency-aware attention modules. Increasingly, these models are also being trained using self-supervised objectives—such as contrastive learning and temporal masking—to enhance their generalization without relying on annotated labels.

Despite these advancements, several limitations remain. Interpretability is still a challenge, especially for models deployed in safety-critical domains like healthcare. Scalability to edge devices is another pressing concern, given the computational demands of ViTs. Furthermore, cross-domain generalization remains difficult, particularly when transitioning between synthetic and real-world datasets or between distinct signal modalities.

Looking ahead, future research will likely focus on developing signal-specific foundation models, designing more efficient and lightweight architectures, and improving self-supervised techniques that can learn from massive unlabeled da-tasets. While this review spans four distinct signal domains to provide a broad foundational perspective, this scope naturally limits the depth of domain-specific technical analysis. Future work may explore each domain in greater detail to uncover specialized modeling strategies and performance nuances. The integration of domain knowledge into model design and pretraining strategies may also play a key role in enhancing model performance and interpretability.

Overall, AEs and ViTs—along with their hybrid forms—are reshaping how time-series signals are processed and understood. Their growing adoption across disciplines reflects not only their versatility but also their promise in addressing the growing demands of data-driven signal intelligence in real-world systems.

## Conflicts of Interest

The authors declare no conflicts of interest regarding the publication of this paper.